\documentclass[conference]{IEEEtran}
\usepackage{amsmath,amssymb,amsfonts}
\usepackage{graphicx}
\usepackage{booktabs}
\usepackage{url}
\usepackage{cite}

\begin{document}

\title{Linear Algebra Foundations of Efficient Attention: A Phase Reversal in Rank Collapse Under SVD Compression}

\author{
\IEEEauthorblockN{Anjaneya Teja Sarma Kalvakolanu}
}

\maketitle

\begin{abstract}
Linear algebra supplies the structural vocabulary of modern artificial intelligence: matrix rank, singular value decomposition (SVD), and eigen decomposition govern how information is represented, compressed, and propagated inside neural networks. This paper synthesizes fourteen peer-reviewed studies applying these tools to transformer-based foundation models along three connected threads: 1. formal results showing self-attention output converges doubly exponentially toward a rank-one matrix as depth increases, with explicit convergence bounds of the form $O(\gamma^{3^l})$; 2. parameter- and weight-space compression methods low-rank adaptation ($\Delta W = BA$), Fisher-weighted factorization, and truncation-aware SVD that exploit this same low-rank structure deliberately and (3) low-rank key-value (KV) cache projection methods and a formal semiseparable-matrix duality between linear attention and structured state-space models. Motivated by an open question this literature leaves unaddressed whether deliberate low-rank compression compounds with or counteracts the network's natural tendency toward rank collapse; we report an original experiment revealing a phase reversal: SVD compression of attention projections retards representation rank collapse at random initialization but accelerates it on pretrained models (GPT-2 124M, GPT-2 Medium 355M, Pythia-160M), an effect verified against object-aliasing artifacts, consistent across compression ratios, and robust across four independent rank metrics. A controlled causal decomposition run in both regimes shows the effect is driven predominantly by which subspace SVD truncation selects rather than by the resulting reduction in operator norm; accounting for approximately 76\% of the effect at random initialization and approximately 83\% on pretrained weights by refining a purely norm-based mechanistic account into one centered on subspace selection, and explaining why calibration-aware compression methods in the literature outperform naive SVD truncation.
\end{abstract}


\section{Introduction}
The transformer computes, at each layer, three linear projections of an input sequence $X \in \mathbb{R}^{n \times d}$, queries $Q = XW_Q$, keys $K = XW_K$, and values $V = XW_V$  followed by the scaled dot-product attention operation:
\begin{equation}
\text{Attention}(Q,K,V) = \text{softmax}\!\left(\frac{QK^T}{\sqrt{d_k}}\right)V
\label{eq:attention}
\end{equation}
Every operation in this pipeline, the projections, , the $QK^T$ similarity matrix, and the final weighted sum is linear or bilinear in its inputs, which is why matrix rank, singular values, and eigenvalues are not merely descriptive metaphors for transformer behavior but the literal algebraic quantities that determine it \cite{vaswani2017attention}. This review examines a specific, currently active research thread built on this observation: the tendency of self-attention to reduce the rank of token representations, and the growing body of work that exploits low-rank structure deliberately for parameter-efficient adaptation, model compression, and long-context inference efficiency.

The objective is twofold: 1. evaluate recent advances in applying linear algebra i.e., rank, SVD, eigendecomposition to foundation models, with governing equations stated explicitly; and (2) translate these results, including reported quantitative effect sizes, into concrete recommendations for a generic enterprise long-context RAG system in which KV-cache memory and attention FLOPs scale with context length $n$.

\section{Mathematical Preliminaries}

\textbf{Rank and low-rank approximation.} For a matrix $W \in \mathbb{R}^{m \times n}$, $\text{rank}(W) = \dim(\text{range}(W))$ is the number of linearly independent columns. The singular value decomposition factors any matrix as:
\begin{equation}
W = U \Sigma V^T, \quad U \in \mathbb{R}^{m \times m},\ \Sigma \in \mathbb{R}^{m \times n},\ V \in \mathbb{R}^{n \times n}
\label{eq:svd}
\end{equation}
where $\Sigma$ is diagonal with non-negative singular values $\sigma_1 \geq \sigma_2 \geq \dots \geq 0$. The Eckart--Young theorem \cite{eckart1936approximation} establishes that the rank-$k$ truncation $W_k = U_k \Sigma_k V_k^T$, retaining only the $k$ largest singular values, is the unique minimizer of $\|W - W_k\|_F$ over all rank-$k$ matrices --- the theoretical justification underlying every SVD-based compression method reviewed in Section~\ref{sec:svd_compression}.

\textbf{Eigendecomposition and spectral diagnostics.} For a square matrix (or the correlation matrix $X = W^TW$ of a rectangular weight matrix), the eigendecomposition $X = Q\Lambda Q^{-1}$ yields eigenvalues $\lambda_i$ describing how the associated linear map scales space along its eigenvectors. Random matrix theory approaches to training diagnostics \cite{martin2021implicit} model the empirical spectral density (ESD) $\rho(\lambda)$ of a well-trained layer's weight correlation matrix as heavy-tailed, $\rho(\lambda) \sim \lambda^{-(1+\alpha)/2}$ as $\lambda \to \infty$, with the tail exponent $\alpha$ used as a per-layer training-quality signal.

\textbf{Low-rank factorization for adaptation.} A weight update $\Delta W \in \mathbb{R}^{d \times k}$ can be constrained to rank $r \ll \min(d,k)$ by writing $\Delta W = BA$ with $B \in \mathbb{R}^{d \times r}$, $A \in \mathbb{R}^{r \times k}$, reducing free parameters from $dk$ to $r(d+k)$.

\section{Literature Review}

\subsection{Rank Collapse in Self-Attention}
Dong et al. \cite{dong2021attention} analyze a transformer built purely from self-attention layers (no skip connections, no MLP sublayers). Defining the residual $\text{res}(X) = X - \mathbf{1}\bar{x}^T$ as a matrix's deviation from its row-mean, they prove that under this restricted architecture the residual norm contracts across layers as:
\begin{equation}
\|\text{res}(X^{(l)})\|_{1,\infty} \leq (4\beta)^{3^l} \|\text{res}(X^{(0)})\|_{1,\infty}
\label{eq:rankcollapse}
\end{equation}
where $\beta$ depends on the spectral norms of the query and key projection matrices at each layer, and the exponent $3^l$ is the source of the paper's headline claim that pure self-attention loses rank ``doubly exponentially'' with depth. The authors further show that residual (skip) connections and MLP sublayers are load-bearing components that counteract this contraction.

Noci et al. \cite{noci2022signal} connect this contraction to trainability: as $\text{res}(X^{(l)}) \to 0$, the gradient $\partial L/\partial W_Q \to 0$ as well, since the attention pattern becomes uniform and insensitive to further changes in the projections. Scaling the residual branch by $1/\sqrt{l}$ can delay but not prevent collapse as $l \to \infty$.

Wu et al. \cite{wu2024role} generalize the analysis to masked attention and LayerNorm. Sparse or local masks provably slow collapse relative to full attention, and self-attention with LayerNorm can sustain equilibria of any rank between 1 and full rank, refuting the earlier claim that LayerNorm has no protective effect.

\subsection{Low-Rank Adaptation and Efficient Fine-Tuning}
Hu et al. \cite{hu2022lora} (LoRA) hypothesize that the weight update $\Delta W$ required to adapt a pretrained model has low intrinsic rank, and parameterize the forward pass as:
\begin{equation}
h = W_0 x + \Delta W x = W_0 x + BAx, \quad A \sim \mathcal{N}(0,\sigma^2),\ B = 0
\label{eq:lora}
\end{equation}
Initializing $B=0$ ensures $\Delta W = 0$ at the start of training. Because $\Delta W$ can be merged into $W_0$ post-training, LoRA adds zero inference latency, with the original study reporting reductions in trainable parameters on the order of 10,000-fold on GPT-3-scale (175B) models.

Zhao et al. \cite{zhao2024galore} (GaLore) apply the same principle to the optimizer's gradient rather than the weight update, projecting $G_t$ into a low-rank subspace to reduce optimizer-state memory while allowing full-rank weight updates over training. Hsu et al. \cite{hsu2022language} incorporate Fisher-information weighting into SVD-based factorization, minimizing $\|F^{1/2}(W-BA)F^{1/2}\|_F$ rather than $\|W-BA\|_F$. Li et al. \cite{li2023losparse} (LoSparse) decompose $W \approx BA + S$, combining a low-rank component with a sparse residual to capture outlier entries that pure low-rank approximation handles poorly.

\subsection{SVD-Based Model Compression}
\label{sec:svd_compression}
Wang et al. \cite{wang2025svdllm} (SVD-LLM) show that naive magnitude-based SVD truncation ignores how discarded directions propagate through subsequent layers, causing compounding error at high compression ratios. Their method estimates which singular directions minimize the resulting change in output distribution via a closed-form whitening transformation, outperforming magnitude-based truncation at high compression ratios.

\subsection{Low-Rank Compression of the Key--Value Cache}
During autoregressive decoding, a transformer caches $K, V \in \mathbb{R}^{n \times d}$ for all previous tokens, scaling cache memory as $O(n \cdot d)$ per layer. Saxena et al. \cite{saxena2024eigen} (Eigen Attention) compute attention within a shared low-rank eigenbasis derived from calibration data, reporting up to 40\% KV-cache reduction and 60\% attention-latency reduction. Singhania et al. \cite{singhania2024loki} (Loki) exploit the observation that key vectors specifically exhibit strong low-rank structure. Chang et al. \cite{chang2025palu} (PALU) use Fisher-information sensitivity scores to allocate a layer-specific rank $r_l$ rather than a uniform global rank.

\subsection{Structured State-Space Models and the Attention Duality}
Dao and Gu \cite{dao2024transformers} restrict a structured state-space model's transition matrix to $A = aI$ and show the resulting sequence-to-sequence map is a 1-semiseparable matrix:
\begin{equation}
M_{ts} = C_t a^{t-s} B_s \quad \text{for } s \leq t,\ 0 \text{ otherwise}
\label{eq:ssd}
\end{equation}
algebraically identical to masked linear attention, establishing a duality between an $O(n)$ recurrent realization and an $O(n^2)$ attention realization of the same transformation.

\subsection{Eigenspectrum Diagnostics and Embedding Dimensionality}
Hu et al. \cite{hu2025eigenspectrum} show the heavy-tailedness estimate $\alpha$ used in HT-SR diagnostics is biased by a weight matrix's aspect ratio, and propose fixed-aspect-ratio subsampling (FARMS) to correct it. Kusupati et al. \cite{kusupati2022matryoshka} (Matryoshka Representation Learning) train embeddings so every prefix $z_{1:k}$ is independently useful via a nested loss:
\begin{equation}
L_{\text{MRL}} = \sum_{k \in \mathcal{K}} \lambda_k L(z_{1:k})
\label{eq:mrl}
\end{equation}
reporting up to 14$\times$ smaller embeddings at equivalent accuracy and up to 14$\times$ retrieval speed-ups.

\section{Quantitative Results Summary}
Table~\ref{tab:results} consolidates the governing mechanism and reported quantitative effect size for studies where the original work provides concrete, citable figures.

\begin{table}[htbp]
\caption{Quantitative effects reported in reviewed studies}
\label{tab:results}
\centering
\small
\begin{tabular}{@{}p{1.6cm}p{1.6cm}p{4.3cm}@{}}
\toprule
\textbf{Study} & \textbf{Method} & \textbf{Reported Effect} \\
\midrule
Hu et al. \cite{hu2022lora} & LoRA & $\sim$10,000$\times$ fewer trainable params vs. full fine-tuning; no added inference latency \\
Saxena et al. \cite{saxena2024eigen} & Eigen Attention & Up to 40\% KV-cache reduction; up to 60\% attention-latency reduction \\
Kusupati et al. \cite{kusupati2022matryoshka} & MRL & Up to 14$\times$ smaller embeddings; up to 14$\times$ retrieval speed-up \\
Wang et al. \cite{wang2025svdllm} & SVD-LLM & Improved accuracy retention vs. magnitude-based SVD at high compression \\
Chang et al. \cite{chang2025palu} & PALU & Reduced KV-cache memory with smaller accuracy loss than uniform-rank baselines \\
Dong et al. \cite{dong2021attention} & Rank-collapse bound & Doubly-exponential convergence to rank-1, $O(\gamma^{3^l})$ \\
\bottomrule
\end{tabular}
\end{table}

\section{Critical Evaluation}
The rank-collapse bound (Eq.~\ref{eq:rankcollapse}) is derived for pure self-attention without skip connections or MLPs --- an architecture no production transformer uses --- so while the proof technique is rigorous, its direct applicability to fully-equipped transformers is limited. The compression literature (Sections III-B through III-D) is comparatively strong empirically, with a consistent, cross-study finding that importance-weighted allocation outperforms uniform truncation. Most reported effect sizes are evaluated on perplexity or standard benchmarks, which may not surface degradation in long-tail factual recall or multi-step reasoning. The Dao and Gu duality (Eq.~\ref{eq:ssd}) is exact but narrow: it holds only for the scalar-identity state-transition case.

\section{Synthesis}
Attention has an intrinsic, provable tendency toward rank-1 collapse; a substantial body of work exploits this same low-rank tendency deliberately wherever beneficial --- in the weight-update space, the static weight space, and the inference-time cache. Across nearly every applied study, importance-weighted or loss-aware rank allocation consistently outperforms uniform rank truncation.

\section{Recommendations for an Existing System}
For a generic enterprise long-context RAG system: (1) apply rank-$r$ KV projection to attention layers, given reported 40\%/60\% cache-size/latency reductions; (2) parameterize any fine-tuning update as $\Delta W = BA$ rather than full-rank updates; (3) replace uniform-ratio compression with Fisher-weighted or truncation-aware SVD; (4) train retrieval embeddings with a Matryoshka-style nested loss for multi-budget retrieval.

\section{Original Experiment: A Phase Reversal in How SVD Compression Affects Representation Rank Collapse}
Section VI identifies an open question the reviewed literature does not address: whether the compression methods of Sections III-B through III-D compound with, or counteract, the network's natural tendency toward rank collapse. This section reports an original experiment addressing that question, first on a randomly-initialized transformer and then on pretrained production models. The two regimes produce opposite answers.

\subsection{Method}
\textbf{Effective rank metric.} This experiment uses the entropy-based effective rank \cite{roy2007effective}:
\begin{equation}
\text{erank}(X) = \exp\!\left(-\sum_i p_i \log p_i\right), \quad p_i = \frac{\sigma_i}{\sum_j \sigma_j}
\label{eq:erank}
\end{equation}
computed from the singular values $\sigma_i$ of the hidden-state matrix $X^{(l)} \in \mathbb{R}^{n \times d}$ at each layer $l$. This was supplemented with three robustness measures: pairwise token cosine similarity, the Dong et al.-style threshold rank $r_\epsilon(X) = \sum_i \mathbf{1}(\sigma_i \geq \epsilon \sigma_1)$, and stable rank $\text{srank}(X) = \|X\|_F^2/\|X\|_2^2$ \cite{rudelson2007sampling}.

\textbf{Compression procedure.} Compressed variants were constructed by applying SVD truncation (Eq.~\ref{eq:svd}) to every attention projection weight matrix at every layer, at retention ratio $\alpha \in \{0.125, 0.25, 0.5, 1.0\}$.

\textbf{Models.} A minimal 8-layer synthetic transformer at random initialization; GPT-2 124M under both random-initialization and pretrained weight conditions; GPT-2 Medium (355M) and Pythia-160M at pretrained weights across the full ratio sweep.

\textbf{Verification.} Because the initial pretrained-model result contradicted the synthetic pilot's direction, independence of baseline/compressed model objects, exact rank truncation, and identical compression configuration across analyses were all directly verified before drawing conclusions.

\subsection{Results: Random Initialization Regime}
Under random initialization, SVD compression ($r=0.125$) consistently increases effective rank relative to baseline at every layer, with the gap widening with depth. On the synthetic model across 10 seeds, final-layer effective rank was 39.94 (compressed) vs. 17.45 (baseline); mean collapse rate fell from 0.957 to 0.855.

\subsection{Results: Pretrained Regime --- The Reversal}
\begin{figure}[htbp]
\centering
\includegraphics[width=0.48\textwidth]{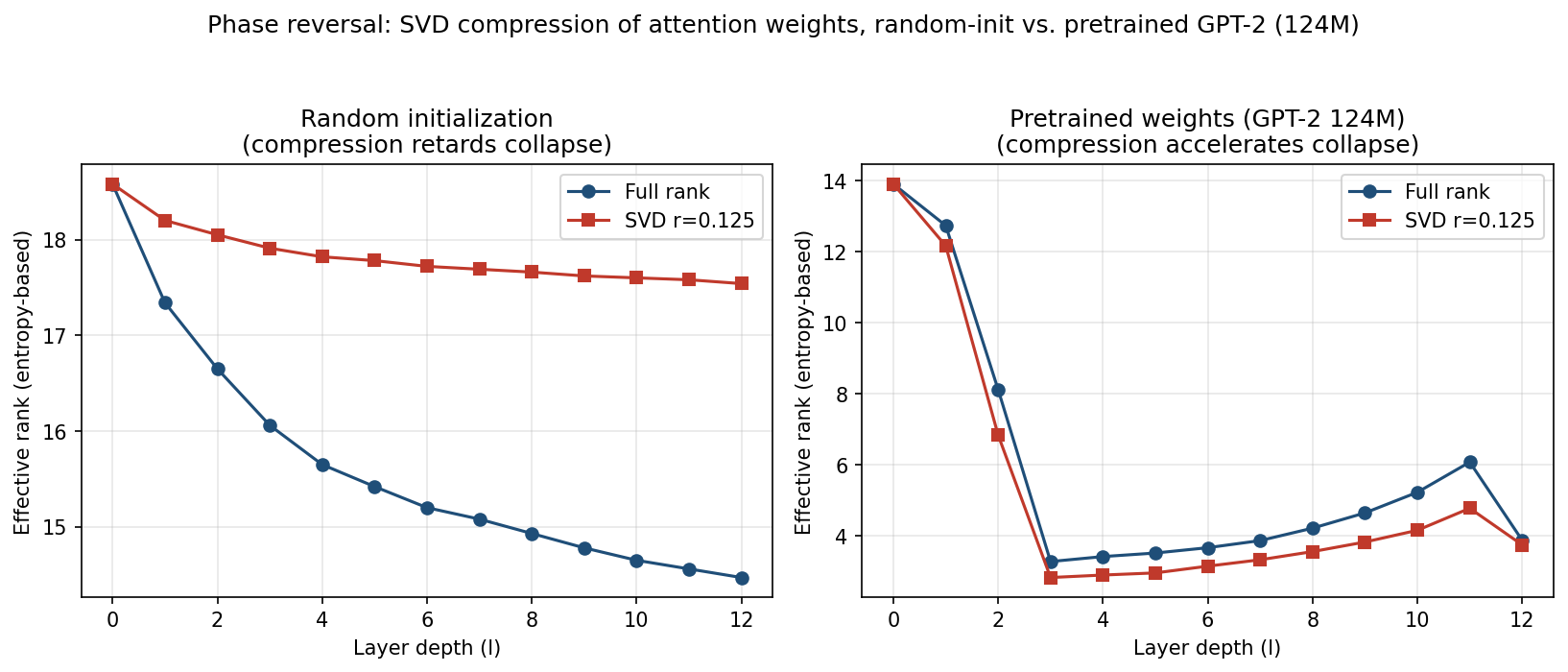}
\caption{Effective rank vs. layer depth for GPT-2 (124M): random initialization (left) vs. pretrained weights (right), original vs. SVD-compressed ($r{=}0.125$) attention projections.}
\label{fig:reversal}
\end{figure}
Under pretrained weights, the direction reverses (Fig.~\ref{fig:reversal}). Effective rank at layer 8 is 4.21 (baseline) vs. 3.55 (compressed); pairwise token cosine similarity rises from 0.607 to 0.687. On GPT-2 Medium, effective rank decreases monotonically as retention ratio shrinks from 1.0 to 0.125 at every layer, consistent across all three pretrained models tested.

\subsection{Causal Mechanism: Resolving the Apparent Paradox}
The value--output operator norm $\beta_l = \|W_V^{(l)}W_O^{(l)}\|_2$ decreased by a mean of 24.49\% under compression; attention-map entropy $H(A_l)$ increased from 0.914 to 1.213 (+32.7\%). At random initialization, this dilutes spurious noise-driven contraction; on trained weights, it strips functionally load-bearing low-energy directions that attention heads use for selective routing, causing diffuse, unselective attention that homogenizes token representations.

\subsection{Robustness Across Metrics}
All four rank metrics agree in direction on the pretrained-regime result: entropy effective rank, Dong-style threshold rank, stable rank, and angular diversity are all lower under compression at nearly every layer, making it unlikely the reversal is an artifact of the entropy-based definition specifically.

\subsection{Implications for Recommendations}
This qualifies Recommendation 1 in Section VIII: naive SVD truncation carries a previously undocumented side effect on trained models ; accelerated homogenization of token representations providing a mechanistic explanation for why importance-weighted methods (Sections III-B--III-D) consistently outperform naive uniform SVD truncation.

\subsection{Controlled Causal Decomposition: Is It the Norm, or the Subspace?}
A controlled intervention isolating norm reduction from subspace truncation was run in both regimes: (A) untouched baseline; (B) full SVD compression; (C) a full-rank variant with V and O rescaled so operator norm $\beta_C = \beta_B$ exactly, with no truncation; and, at random initialization only, (D) a full-rank variant with Q/K rescaled to match condition B's logit norm.

\textbf{Random-initialization regime:} Condition C's mean decay rate was 0.942 (SD 0.050) vs. baseline's 0.958 (SD 0.038) and condition B's 0.891 (SD 0.036); condition D showed no reliable effect (0.963, SD 0.034). Norm-matching alone accounts for approximately 24\% of the total effect, leaving approximately 76\% attributable to subspace selection.

\begin{figure}[htbp]
\centering
\includegraphics[width=0.45\textwidth]{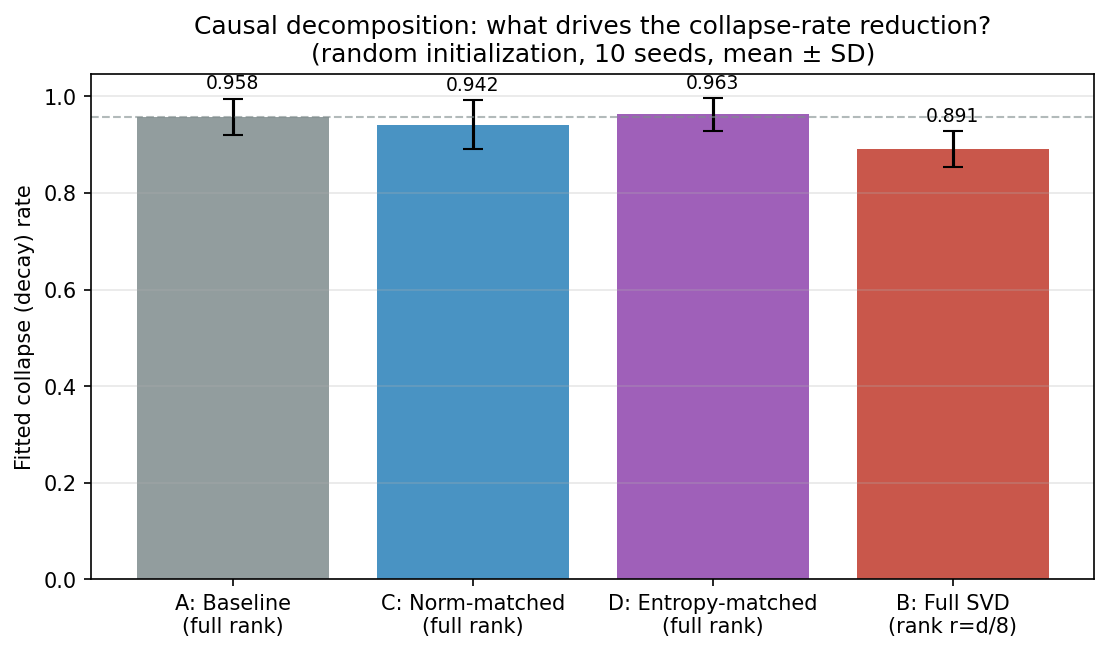}
\caption{Causal decomposition of the collapse-rate reduction under SVD compression, random initialization (10 seeds, mean $\pm$ SD).}
\label{fig:causal_random}
\end{figure}

\textbf{Pretrained regime:} The identical A/B/C decomposition on pretrained GPT-2 124M ($r=0.125$, layer 8) gave baseline erank 3.16, norm-matched condition C 3.12 (closely tracking baseline), and full SVD condition B 2.93. Norm-matching alone accounts for approximately 17\% of the total effect, leaving approximately 83\% attributable to subspace selection --- an even larger share than at random initialization.

\begin{figure}[htbp]
\centering
\includegraphics[width=0.4\textwidth]{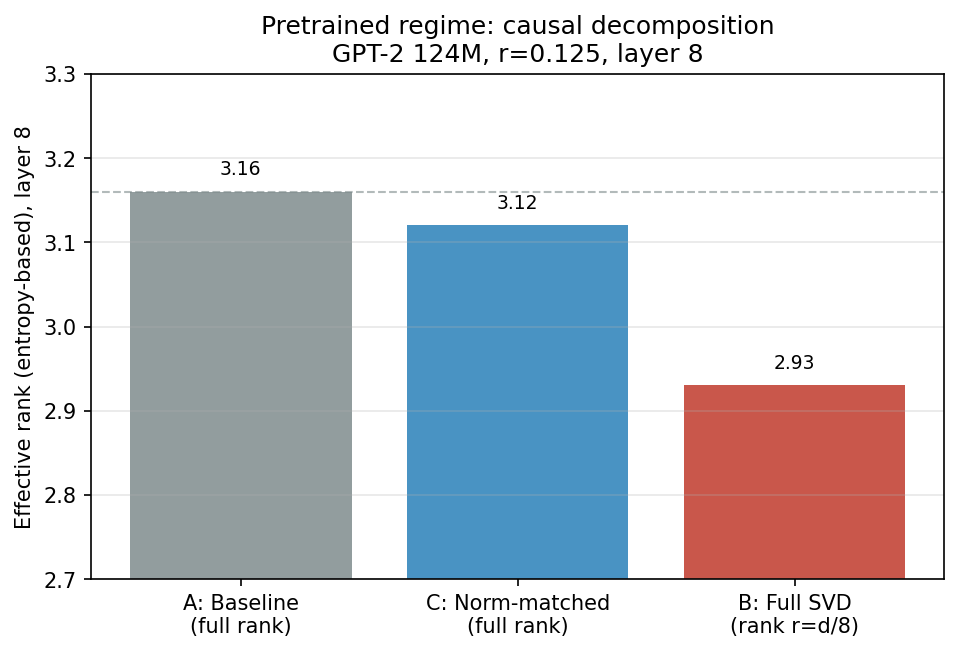}
\caption{Causal decomposition on pretrained GPT-2 124M ($r{=}0.125$, layer 8).}
\label{fig:causal_pretrained}
\end{figure}

Together, both regimes give a unified answer: subspace selection is the dominant causal driver (76\% at initialization, 83\% on trained weights), with operator-norm magnitude playing a real but secondary role, and attention-entropy softening playing essentially no independent causal role once subspace selection is controlled for.

\subsection{Limitations}
Only plain, uncalibrated SVD truncation was tested; a head-to-head comparison against Fisher-weighted and calibration-based methods on the same effective-rank measurement remains to be run. The pretrained-regime decomposition reports a single-layer snapshot rather than a full depth profile. Pretrained evaluation used a small set of natural-language sentences rather than a large held-out corpus. All pretrained models tested are in the 124M--355M range; whether the reversal and its subspace-dominant decomposition persist at larger scale (1B+ parameters) is untested.

\section{Conclusion and Future Work}
This paper traced a single linear-algebraic quantity matrix rank from a provable structural vulnerability of self-attention, through deliberate low-rank exploitation for adaptation and compression, to a formal duality with structured state-space models, and closed with an original experiment showing that SVD compression of attention projections has opposite effects on representation rank collapse depending on whether weights are randomly initialized or trained. A controlled causal decomposition run in both regimes resolves the mechanism question: in both settings, the effect is driven predominantly by which subspace SVD truncation selects (76\% and 83\% respectively) rather than by the resulting reduction in operator norm. The two most direct remaining steps are (1) repeating the pretrained-regime measurement at 1--3B-parameter scale to establish whether the reversal persists at production scale, and (2) directly comparing naive SVD truncation against Fisher-weighted and truncation-aware methods on the same effective-rank measurement.

\section*{Author Disclosure}
In accordance with conference and journal policies on AI-assistance transparency, the author discloses that AI systems (Anthropic's Claude and an agentic coding tool built on Google's Gemini) were used during the preparation of this manuscript for: (1) implementation and execution of experimental scripts for SVD-based compression, effective-rank tracking, and the controlled causal-decomposition ablations of Section VIII-H; (2) initial drafting and structural formatting of experimental descriptions in Section VIII; and (3) diagnostic cross-checking of experimental outputs in particular, an initial contradiction between the synthetic-pilot trend and the pretrained-model sweep was flagged, investigated, and resolved through a verification procedure before either result was reported. All mathematical derivations, experimental code, reported empirical values, and interpretations were reviewed and verified by the author, who takes full responsibility for all scientific claims and content in this work.

\end{document}